\documentclass[runningheads]{llncs}

\usepackage{eccv}

\usepackage{eccvabbrv}

\usepackage{graphicx}
\usepackage{booktabs}

\usepackage[accsupp]{axessibility}  

\usepackage{hyperref}

\usepackage{orcidlink}

\usepackage{tikz}
\usetikzlibrary{arrows.meta}

\begin{document}


\title{HSEmotion Team at the 11th ABAW Challenge:
Multi-Task Learning and Ambivalence/Hesitancy Video Recognition}
\titlerunning{HSEmotion Team at the 11th ABAW Challenge}

\author{Aleksei Bakin\inst{1}\orcidlink{0009-0005-8845-2945} \and Andrey V. Savchenko\inst{2,3}\orcidlink{0000-0001-6196-0564}}

\authorrunning{A.A.~Bakin, A.V.~Savchenko}

\institute{Central University, Moscow, Russia\\ \email{bakin95aa@gmail.com} \and Sber AI Lab, Moscow, Russia \and
HSE University, Laboratory of Algorithms and Technologies for Network Analysis, Nizhny Novgorod, Russia\\
\email{avsavchenko@hse.ru}}

\maketitle

\begin{abstract}
This article presents our results for the 11th Affective Behavior Analysis in-the-Wild (ABAW) competition. For multi-task learning with simultaneous prediction of valence, arousal, facial expressions, and action units on s-Aff-Wild2 dataset, we use frozen lightweight facial extractors, MT-EmotiDDAMFN and MT-EmotiEffNet-B0, with separate heads and systematic post-processing: temporal Gaussian smoothing, per-class expression bias, AffectNet blending, per-AU threshold tuning, and weighted backbone fusion. On the official validation set, our ensemble significantly exceeds the performance of the ConvNeXt baseline. For ambivalence/hesitancy video recognition on the expanded BAH dataset, we extend the audiovisual pipeline to video-level Macro F1 by late fusion of face, HuBERT audio, and RoBERTa text classifiers, temporal aggregation, and a global-text gate. Frame-level Weighted F1 on validation set rises from 0.74 in ABAW-8 to 0.79, while the best public-test video-level Macro F1 reaches 0.73. In both tasks, competitive performance is achieved without fine-tuning heavy backbones. 
These results indicate that systematic prediction calibration and lightweight multimodal fusion can rival substantially heavier end-to-end approaches while offering improved efficiency and deployment flexibility.

\end{abstract}

\section{Introduction}
\label{sec:intro}

Robust affect analysis in unconstrained environments is essential for human-computer interaction, mental health monitoring, and socially aware AI. The Affective Behavior Analysis in-the-wild (ABAW) workshop series~\cite{kollias2020analysing,kollias2021affect,kollias2023abaw2} has become a major benchmark for this problem, with tasks built on Aff-Wild~\cite{zafeiriou2017aff,kollias2019deep} and Aff-Wild2~\cite{kollias2019expression,kollias2022abaw}. Recent editions increasingly emphasize multimodal cues, temporal reasoning, and reliable evaluation under missing labels and class imbalance. At the same time, top-ranking solutions often rely on large ensembles, self-supervised vision transformers, and heavy fine-tuning~\cite{qiu2024learning,zhang2022multi}, which limits their use on mobile or privacy-sensitive platforms~\cite{kharchevnikova2018neural,savchenko2022cvprw}.

The 11th ABAW competition~\cite{kollias2026abaw11} features two challenges. The first is multi-task learning (MTL)~\cite{kollias2019face,kollias2021distribution}, in which each facial frame must be annotated with valence, arousal, one of eight expressions, and twelve action units. This task continues the s-Aff-Wild2 line studied in ABAW-3~\cite{kollias2022abaw}, ABAW-4~\cite{kollias2023abaw} and ABAW-7. The second challenge is Ambivalence/Hesitancy (A/H) recognition on the BAH dataset~\cite{kollias2025bah}. A/H was introduced in ABAW-8~\cite{kollias2025abaw8} with frame-level Weighted F1; ABAW-10~\cite{kollias2026abaw10} continued the task on the same BAH split with \textbf{video-level Macro F1}. ABAW-11 returns to A/H on an expanded release of $1{,}427$ videos and keeps video-level Macro F1.

Our approach to both challenges follows the same principle: keep strong pre-trained extractors frozen, train only lightweight task-specific classifiers, and invest accuracy gains into post-processing and fusion. For MTL, we reimplement the complete pipeline from ABAW-7 in PyTorch, enabling reproducible training and simplified deployment~\cite{savchenko2024leveraging}. We combine MT-EmotiDDAMFN with MT-EmotiEffNet-B0~\cite{savchenko2023hse,zhang2023dual}. For A/H, we combine MT-EmotiEffNet-B0 face descriptors with HuBERT-large audio and RoBERTa-go\_emotions text embeddings, apply late fusion of modality-specific MLPs, aggregate frame scores inside each video, and optionally filter predictions with a global transcript classifier inherited from ABAW-8~\cite{savchenko2025abaw8}. This design allows facial, audio, and text features to be computed locally, while only compact predictions are combined downstream. Thus, unlike previous ABAW submissions that primarily improve representation learning through larger pretrained encoders or additional self-supervised pretraining, we instead investigate how far frozen lightweight representations can be pushed through systematic prediction calibration and modality-specific post-processing.

Our contributions are fourfold:
\begin{itemize}
\item We demonstrate that frozen lightweight affect recognition backbones combined with lightweight task-specific heads remain highly competitive for both multi-task facial affect analysis and multimodal ambivalence recognition.
\item We show that carefully designed validation-time calibration, including temporal smoothing, expression calibration, AU threshold optimization, and backbone fusion, provides larger gains than replacing the backbone with substantially heavier architectures.
\item We present a unified lightweight framework applicable to two different ABAW challenges, achieving state-of-the-art validation/public-test performance while avoiding end-to-end fine-tuning.
\item We release the complete PyTorch implementation to facilitate reproducible affective computing research.
\end{itemize}

The rest of the paper is organized as follows. Section~\ref{sec:related} reviews related work on MTL and A/H recognition. Section~\ref{sec:methods} presents our pipelines for both challenges. Section~\ref{sec:exper} reports validation and public-test results. Section~\ref{sec:concl} concludes the paper.

\section{Related Works}
\label{sec:related}

\subsection{Multi-Task Learning in ABAW Competitions}

The MTL challenge requires simultaneous prediction of valence, arousal, eight expressions, and twelve action units on s-Aff-Wild2~\cite{kollias2022abaw}. Because ABAW-3 participants were not required to train on this split~\cite{kollias2022abaw}, we focus on ABAW-4~\cite{kollias2023abaw} and ABAW-7~\cite{kollias20247th}, which share the official combined metric $P_{MTL}=P_{VA}+P_{EXPR}+P_{AU}$ and are closest to the ABAW-11 protocol.

Early s-Aff-Wild2 solutions mainly explored task interaction and calibration without large backbones. The two-aspect information interaction model~\cite{sun2022two} models relations between facial sign vehicles and emotional messages. SS-MFAR~\cite{gera2022facial} combines ResNet~\cite{he2016resnet} features with expression-specific thresholds estimated via semi-supervised learning. A hybrid CNN--Transformer~\cite{mtl2022fifth} fuses ResNet-18~\cite{he2016resnet} and a spatial transformer and finished fifth on the validation leaderboard. Cross-attentive AU graphs~\cite{nguyen2022affective} capture dependencies among action units, while SMMEmotionNet ensembles~\cite{mtl2022six} aggregate facial embeddings from multiple extractors. Among lightweight approaches, MT-EmotiEffNet~\cite{savchenko2023hse}---an EfficientNet-B0~\cite{tan2019efficientnet} model pre-trained for joint VA/EXPR learning---reached third place without MAE fine-tuning. The top two validation submissions relied on masked-autoencoder (MAE)~\cite{he2022masked} facial priors: an EMMA ensemble of MAE ViT~\cite{he2022masked,dosovitskiy2021vit} and CNN encoders~\cite{li2023affective} took second place, and an MAE+Transformer ensemble with temporal modeling~\cite{zhang2022multi,he2022masked} won the challenge.

ABAW-7 moved evaluation to the private test split and shifted the field toward heavier representation learning. The winning Netease Fuxi AI Lab pipeline~\cite{liu2024affective} uses a progressive strategy: task-specific heads are trained first on a self-supervised MAE~\cite{he2022masked} extractor, then jointly refined with cross-task feature fusion and a temporal convergence module that models expression dynamics across frames. HFUT-MAC1~\cite{shen2024facial} ranked third with a multi-architecture design that blends MAE~\cite{he2022masked}, ResNet~\cite{he2016resnet}, POSTER, and OpenFace AU descriptors through a Transformer encoder and affine feature alignment. SCU ACers~\cite{li2024affective} placed fourth using frozen DINOv2~\cite{oquab2023dinov2} features, task-adaptive query decoding, and an AU-assisted graph convolutional network that transfers AU structure to EXPR and VA prediction. Our ABAW-7 entry~\cite{savchenko2024leveraging} took second place with a contrasting philosophy: frozen MT-EmotiEffNet and MT-EmotiDDAMFN extractors, lightweight PyTorch heads, Gaussian temporal smoothing, per-AU threshold tuning, and weighted backbone blending, without fine-tuning large transformers. For ABAW-11, organizers provide a frozen ConvNeXt baseline~\cite{liu2022convnext,kollias2026abaw11} as a modern convolutional reference point on the same validation protocol.

\subsection{Ambivalence and Hesitancy Recognition}

Ambivalence/hesitancy (A/H) recognition was introduced in ABAW-8~\cite{kollias2025abaw8} on the BAH dataset~\cite{kollias2025bah}, which provides frame-level labels, speech transcripts, and cropped faces from behavioural Q\&A videos. The first edition evaluated frame-level Weighted F1 on the BAH test split~\cite{kollias2025abaw8}. The organizer baseline is a multimodal model with visual, audio, and transcript features fused through temporal convolution and co-attention~\cite{richet2024text,kollias2025abaw8}.

Top frame-level solutions combined pretrained unimodal encoders with simple fusion. The HSEmotion pipeline~\cite{savchenko2025abaw8} extracts facial descriptors from EmotiEffLib~\cite{savchenko2023emotieffnets}, acoustic embeddings from HuBERT~\cite{hsu2021hubert} and wav2vec~2.0~\cite{baevski2020wav2vec}, and text embeddings from RoBERTa~\cite{liu2019roberta} trained on GoEmotions~\cite{demszky2020goemotions}; non-aligned audio and text are interpolated to frame rate, and lightweight MLP classifiers are fused by blending with temporal smoothing. This audiovisual--text design won the challenge. The runner-up, HCAI-VIS (Semantic Matters)~\cite{hallmen2025semantic}, pursued a complementary semantic route: ViT~\cite{dosovitskiy2021vit} visual features with chunk-wise temporal pooling, emotional-speech Wav2Vec~2.0~\cite{baevski2020wav2vec} features, and BERT text representations fused through MLP/LSTM modules, confirming that transcripts are often the strongest single cue for hesitancy.

ABAW-10~\cite{kollias2026abaw10} reformulated the task as \textbf{video-level Macro F1} on an expanded BAH release, requiring a single binary decision per video rather than per-frame labels. The winning VisPBF solution (BROTHER)~\cite{pereira2026brother} trains separate expert networks on visual, audio, and contextual cues, ensembles them with diversity constraints, and adds sequence modeling for temporal aggregation. Fennec (ConflictAwareAH)~\cite{bekhouche2026conflictaware} extracts VideoMAE~\cite{tong2022videomae}, HuBERT~\cite{hsu2021hubert}, and RoBERTa--GoEmotions~\cite{liu2019roberta,demszky2020goemotions} embeddings and explicitly models cross-modal conflict via pairwise embedding differences before attention pooling. LEYA~\cite{ryumina2026leya} ranked third with a four-stream design---scene VideoMAE~\cite{tong2022videomae}, face EfficientNet-B0~\cite{tan2019efficientnet}, EmotionWav2Vec~2.0 with Mamba temporal encoding, and fine-tuned text transformers---fused through prototype-augmented Transformer modules and ensembled for the final video decision.

ABAW-11 continues video-level Macro F1 on an expanded BAH corpus of $1{,}427$ videos~\cite{kollias2025bah,kollias2026abaw11}. The official zero-shot Video-LLaVA baseline~\cite{lin2024videollava,kollias2026abaw11} remains weak on the public split.

\section{Methods}\label{sec:methods}

We address two tasks of the ABAW-11 challenge. Below we formulate each task and describe the proposed pipeline.

\subsection{MTL Challenge}
\label{sec:methods_mtl}

The MTL challenge requires recognizing the emotions of each video frame $X(t)$, $t=t_1,t_2,\ldots,t_N$, where $1\le t_1<t_2<\ldots<t_N$ are the observed frame indices from s-Aff-Wild2~\cite{kollias2022abaw}. Human affect can be represented both continuously and discretely. In the former case, the most typical emotional space is the two-factor Russell's circumplex model of affect~\cite{russell1980circumplex} with VA-based encoding. Discrete representations include the set of basic expressions of Paul Ekman and the Facial Action Coding System (FACS)~\cite{ekman1978facial} with specific facial action units (AUs).

For the MTL competition, it is necessary to assign $X(t)$ to three emotional representations:
\begin{enumerate}
 \item Valence $V(t) \in [-1,1]$ and arousal $A(t) \in [-1,1]$ (multi-output regression task).
 \item Facial expression $c(t) \in \{1,\ldots,C_{EXPR}\}$, where $C_{EXPR}=8$ is the total number of basic emotions: Neutral, Anger, Disgust, Fear, Happiness, Sadness, Surprise, and Other (multi-class classification).
 \item AUs $\mathbf{AU}(t)=[AU_1(t),\ldots,AU_{C_{AU}}(t)]$, where $C_{AU}=12$ is the total number of AUs and $AU_i(t)\in\{0,1\}$ (multi-label classification).
\end{enumerate}
The official performance measure is $P_{MTL}=P_{VA}+P_{EXPR}+P_{AU}$, where $P_{VA}=(CCC_V+CCC_A)/2$ is the mean Concordance Correlation Coefficient (CCC)~\cite{lawrence1989concordance} of valence and arousal; $P_{EXPR}$ is the macro-averaged F1-score across all eight expression categories; and $P_{AU}$ is the average F1-score across all twelve AUs.

\begin{figure}[t]
\centering
\resizebox{0.92\linewidth}{!}{
\usetikzlibrary{arrows.meta}
\begin{tikzpicture}[
  box/.style={
    draw, rounded corners=2pt, minimum height=0.72cm,
    align=center, font=\scriptsize, inner sep=4pt
  },
  stage/.style={box, fill=gray!10},
  outbox/.style={box, fill=green!12},
  arr/.style={-, semithick}
]

\draw[line width=0.85pt, rounded corners=3pt, fill=white]
  (3.5, -1.8) rectangle (7.1, 1.4);
\draw[line width=0.85pt, rounded corners=3pt, fill=white]
  (3.3, -1.6) rectangle (6.9, 1.6);

\node[box, fill=blue!8, text width=0.9cm] (frames) at (0, 0.05)
  {s-Aff-Wild2\\frame $X(t)$};

\node[stage, text width=1.32cm] (ddamfn) at (1.85, 0.82)
  {Frozen\\MT-Emoti-\\DDAMFN};
\node[stage, text width=1.32cm] (effnet) at (1.85, -0.82)
  {Frozen\\MT-Emoti-\\EffNet-B0};

\node[stage, text width=0.82cm] (h_va) at (4.05, 1.12) {VA\\head};
\node[stage, text width=1.4cm] (p_va) at (5.85, 1.12) {Temporal\\smoothing};

\node[stage, text width=0.82cm] (h_ex) at (4.05, 0.00) {EXPR\\head};
\node[stage, text width=1.4cm] (p_ex) at (5.85, 0.00) {Temporal\\smoothing,\\Expression\\ bias};

\node[stage, text width=0.82cm] (h_au) at (4.05, -1.12) {AU\\head};
\node[stage, text width=1.4cm] (p_au) at (5.85, -1.12) {Threshold\\tuning};

\node[stage, text width=1.10cm] (fuse_va) at (8.15, 1.12) {Backbone\\blending};
\node[outbox, text width=0.50cm] (out_va) at (9.35, 1.12) {$\hat{V},\hat{A}$};

\node[stage, text width=1.10cm] (fuse_ex) at (8.15, 0.0) {Backbone\\blending,\\AffectNet\\blend};
\node[outbox, text width=0.50cm] (out_ex) at (9.35, 0.0) {$\hat{c}$};

\node[stage, text width=1.10cm] (fuse_au) at (8.15, -1.12) {Backbone\\blending};
\node[outbox, text width=0.50cm] (out_au) at (9.35, -1.12) {$\hat{AU}$};

\draw[arr] (frames.east) -- ++(0.25, 0) coordinate (split);
\draw[arr] (split) |- (ddamfn.west);
\draw[arr] (split) |- (effnet.west);

\draw[arr] (ddamfn.east) -| (2.85, 0.82);
\draw[arr, dashed] (effnet.east) -| (3.15, -0.82);

\draw[arr] (2.85, 0.82) |- (h_va.west);
\draw[arr] (2.85, 0.82) |- (h_ex.west);
\draw[arr] (2.85, 0.82) |- (h_au.west);

\draw[arr, dashed] (3.15, -0.82) |- (3.3, 0.92);
\draw[arr, dashed] (3.15, -0.82) |- (3.3, -0.2);
\draw[arr, dashed] (3.15, -0.82) |- (3.3, -1.32);

\draw[arr] (h_va.east) -- (p_va.west);
\draw[arr] (h_ex.east) -- (p_ex.west);
\draw[arr] (h_au.east) -- (p_au.west);

\draw[arr] (p_va.east) -- (fuse_va.west);
\draw[arr] (p_ex.east) -- (fuse_ex.west);
\draw[arr] (p_au.east) -- (fuse_au.west);
\draw[arr] (fuse_va.east) -- (out_va.west);
\draw[arr] (fuse_ex.east) -- (out_ex.west);
\draw[arr] (fuse_au.east) -- (out_au.west);

\draw[arr, dashed] (7.1, 0.92) -- (7.4, 0.92);
\draw[arr, dashed] (7.1, -0.2) -- (7.4, -0.2);
\draw[arr, dashed] (7.1, -1.32) -- (7.4, -1.32);

\end{tikzpicture}}
\caption{Proposed MTL pipeline: two frozen extractors, three task-specific heads, separate post-processing per task, per-task weighted backbone blending, and AffectNet blending for EXPR after backbone blending.}
\label{fig:mtl_pipeline}
\end{figure}
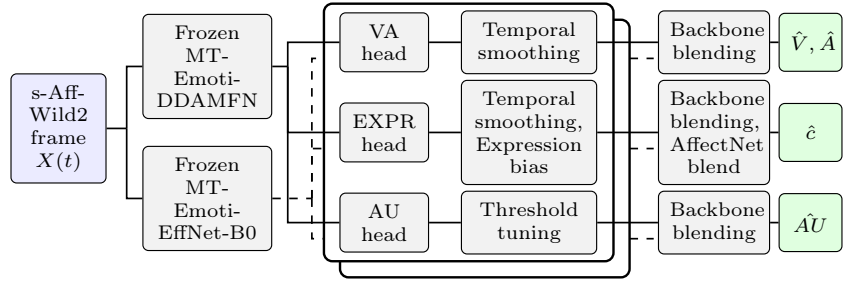

This paper proposes a pipeline for the ABAW-11 MTL challenge (Fig.~\ref{fig:mtl_pipeline}). Compared to our ABAW-7 solution~\cite{savchenko2024leveraging}, we migrate the training code to PyTorch, keep the backbone weights frozen, and add systematic post-processing comprising: (i)~temporal Gaussian smoothing of VA and EXPR predictions; (ii)~per-class expression logit bias; (iii)~per-AU threshold tuning; (iv)~weighted blending of the two backbones; and (v)~AffectNet blending of the fused EXPR predictions.

The main part of the pipeline is the feature extractor backbones based on lightweight neural network architectures~\cite{savchenko2022cvprw,savchenko2023hse,savchenko2024leveraging}. We use MT-EmotiDDAMFN (Dual-Direction Attention Mixed Feature Network)~\cite{zhang2023dual} and MT-EmotiEffNet-B0 (EfficientNet-B0)~\cite{tan2019efficientnet}. Both models were firstly pre-trained to recognize faces from the VGGFace2 dataset~\cite{cao2018vggface2} and next fine-tuned on AffectNet~\cite{mollahosseini2017affectnet} to simultaneously classify static facial expressions and predict VA using a multi-task loss~\cite{savchenko2023hse}, which is essentially a sum of weighted categorical cross-entropy for facial expressions and CCC for valence and arousal. During MTL training on s-Aff-Wild2, the backbone weights remain frozen. We hypothesize that AffectNet-pretrained facial representations already capture most low-level affective information required by ABAW. Consequently, optimizing lightweight prediction heads and prediction calibration becomes more beneficial than updating backbone parameters, particularly under relatively limited challenge training data.

MT-EmotiDDAMFN produces a $522$-dimensional descriptor ($512$-dimensional embedding $\mathbf{x}$ plus $10$-dimensional scores $\mathbf{s}$), while MT-EmotiEffNet-B0 produces a $1290$-dimensional descriptor ($1280$-dimensional embedding $\mathbf{x}$ plus $10$-dimensional scores $\mathbf{s}$). In both cases, $\mathbf{s}$ contains eight logits of AffectNet expressions plus valence and arousal predictions at the output of the last layer.

Next, we train lightweight PyTorch heads on the official \texttt{cropped\_aligned} subset of s-Aff-Wild2. The EXPR head is a linear layer with softmax outputs $\mathbf{p}_{EXPR}\in[0,1]^8$. The VA head uses only the ten AffectNet scores $\mathbf{s}$ (similarly to the slice layer in~\cite{savchenko2023hse}) and returns $\mathbf{p}_{VA}\in[-1,1]^2$ with hyperbolic tangent activations. The AU head is a feed-forward network with one hidden layer ($128$ units, ReLU) and $12$ logistic sigmoid outputs $\mathbf{p}_{AU}\in[0,1]^{12}$. The inputs of the EXPR and AU heads are the concatenation $[\mathbf{x},\mathbf{s}]$. Each head is trained separately with early stopping on its validation metric (macro F1 for EXPR and AU, mean CCC for VA).

As sequential emotions should be smooth, we apply task-specific post-processing inside each video, fuse the two backbones with task-specific weights, and finally apply AffectNet blending to the fused EXPR stream (Fig.~\ref{fig:mtl_pipeline}). The steps are:

\begin{enumerate}

\item \textbf{Temporal smoothing.} Dynamic changes of facial expressions and VA may be significantly different, so we use separate parameters $\sigma_{EXPR}$, $\sigma_{VA}$, $\delta_{EXPR}$, and $\delta_{VA}$. In practice, smoothing decreases the quality of AU detection, so AU predictions are not smoothed. For VA and EXPR, frame-wise predictions of each backbone are smoothed inside each video with a Gaussian kernel:
\begin{equation}\label{eq:gauss}
\overline{\mathbf{p}}(t)=\frac{\sum \limits_{t_i \in \mathcal{N}_\delta(t)}{\exp\!\left(-\frac{(t_i-t)^2}{\sigma}\right)\mathbf{p}(t_i)}}{\sum \limits_{t_i \in \mathcal{N}_\delta(t)}{\exp\!\left(-\frac{(t_i-t)^2}{\sigma}\right)}},
\end{equation}
 where $\mathcal{N}_\delta(t)$ contains frames from the same video within $\delta$ frames of $t$. The parameters $\sigma$ and $\delta$ are selected on the validation set for each backbone and task separately.

 \item \textbf{Expression bias.} For each backbone, a per-class logit bias vector $\mathbf{b}\in\mathbb{R}^{8}$ for EXPR is estimated by coordinate search on the validation set. The biased EXPR label of backbone $b$ is defined as
\begin{equation}\label{eq:expr_bias}
\hat{c}^{(b)}_{MTL}(t)=\arg\max_{k}\left(\log(\overline{p}^{(b)}_{EXPR,k}(t)+\epsilon)+b_k\right),
\end{equation}
 where $\overline{\mathbf{p}}^{(b)}_{EXPR}(t)$ are the smoothed EXPR probabilities of backbone $b$ and $\epsilon$ is a small constant for numerical stability.

 \item \textbf{AU threshold tuning.} Each AU is predicted independently by comparing the head score with a class-specific threshold:
\begin{equation}\label{eq:au_thresh}
AU_i(t)=\begin{cases}
1, & \text{if } p_{AU,i}(t)\ge t_{AU,i},\\
0, & \text{otherwise},
\end{cases}
\quad i=1,\ldots,12.
\end{equation}
 We consider a fixed threshold $t_{AU,i}=0.5$ for all AUs and the best thresholds $t^*_{AU,i}$ that maximize the mean F1-score over twelve AUs on the validation set.

 \item \textbf{Backbone blending.} Frame-wise predictions of the two backbones are fused with task-specific weights $w_{VA}$, $w_{EXPR}$, and $w_{AU}$ chosen on the validation set:
\begin{align}\label{eq:blending}
\mathbf{p}^{(blend)}_{VA}(t)&=w_{VA}\cdot\overline{\mathbf{p}}^{(1)}_{VA}(t) + (1-w_{VA})\cdot\overline{\mathbf{p}}^{(2)}_{VA}(t), \\
\mathbf{p}^{(blend)}_{EXPR}(t)&=w_{EXPR}\cdot\mathbf{p}^{(1)}_{EXPR,bias}(t) + (1-w_{EXPR})\cdot\mathbf{p}^{(2)}_{EXPR,bias}(t), \\
\mathbf{p}^{(blend)}_{AU}(t)&=w_{AU}\cdot\mathbf{p}^{(1)}_{AU}(t) + (1-w_{AU})\cdot\mathbf{p}^{(2)}_{AU}(t),
\end{align}
 where $\mathbf{p}^{(b)}_{EXPR,bias}(t)$ is the one-hot (or soft) EXPR distribution induced by $\hat{c}^{(b)}_{MTL}(t)$, and $\mathbf{p}^{(b)}_{AU}(t)$ are raw AU scores before thresholding. Our best configuration is $w_{VA}=1.0$, $w_{EXPR}=0.4$, and $w_{AU}=0.5$, meaning that DDAMFN dominates VA estimation, while EXPR and AU predictions are mixed almost equally.

 \item \textbf{AffectNet blending.} After backbone fusion, let $p^{AffectNet}_k(t)$ be the softmax probability of the $k$-th AffectNet expression computed from the frozen scores $\mathbf{s}(t)$ of MT-EmotiDDAMFN. The most confident AffectNet label among the first seven basic expressions is
\begin{equation}\label{eq:affectnet_argmax}
\hat{k}(t)=\arg\max_{k<7}p^{AffectNet}_k(t).
\end{equation}
 The fused MTL label is $\hat{c}_{MTL}(t)=\arg\max_k p^{(blend)}_{EXPR,k}(t)$. The final EXPR label is
\begin{equation}\label{eq:expr_blend}
\hat{c}_{EXPR}(t)=
\begin{cases}
\hat{k}(t), & \text{if } p^{AffectNet}_{\hat{k}(t)}(t)>\tau_{blend},\\
\hat{c}_{MTL}(t), & \text{otherwise},
\end{cases}
\end{equation}
 where $\tau_{blend}=0.88$ for the final ensemble. For single-backbone ablations, we use $\tau_{blend}=0.85$ for MT-EmotiDDAMFN and $\tau_{blend}=0.90$ for MT-EmotiEffNet-B0.
\end{enumerate}

The final VA prediction is $\mathbf{p}^{(blend)}_{VA}(t)$. The EXPR output is $\hat{c}_{EXPR}(t)$ from Eq.~(\ref{eq:expr_blend}). For AU detection, thresholds $t^*_{AU,i}$ are tuned on $\mathbf{p}^{(blend)}_{AU}(t)$ and applied according to Eq.~(\ref{eq:au_thresh}).

\subsection{A/H Video Recognition Challenge}
\label{sec:methods_ah}

The A/H challenge is a \textbf{binary video-level classification} problem. For each video $V$ from the BAH dataset~\cite{kollias2025bah}, the goal is to predict whether ambivalence or hesitancy is present ($y=1$) or absent ($y=0$). The expanded ABAW-11 release contains $1{,}427$ videos ($10.6$ hours, $300$ participants) with frame- and video-level expert annotations, speech transcripts, and cropped-aligned faces. The official metric is \textbf{video-level Macro F1} across both classes; we also report average precision (AP) of the positive class.

We extend ABAW-8 audiovisual pipeline~\cite{savchenko2025abaw8} from frame-level to video-level prediction. The pipeline extracts frozen multimodal descriptors, trains lightweight frame classifiers, fuses their outputs, aggregates scores inside each video, and optionally applies a global text gate.

\begin{figure}[t]
\centering
\resizebox{0.92\linewidth}{!}{
\usetikzlibrary{arrows.meta}
\begin{tikzpicture}[
  box/.style={
    draw, rounded corners=2pt, minimum height=0.72cm,
    align=center, font=\scriptsize, inner sep=4pt
  },
  stage/.style={box, fill=gray!10},
  outbox/.style={box, fill=green!12},
  arr/.style={-, semithick}
]

\node[box, fill=blue!8, text width=1.15cm] (video) at (0, 0)
  {BAH video\\$V$, frame $t$};

\node[stage, text width=1.42cm] (face) at (1.95, 1.32)
  {Frozen\\MT-Emoti-\\EffNet-B0};
\node[stage, text width=1.42cm] (audio) at (1.95, 0.00)
  {Frozen\\HuBERT-\\large};
\node[stage, text width=1.42cm] (text) at (1.95, -1.32)
  {Frozen\\RoBERTa-\\GoEmotions};

\node[stage, text width=0.8cm] (interp_a) at (3.6, 0.00) {Interp\\to frames};
\node[stage, text width=0.8cm] (interp_t) at (3.6, -1.32) {Interp\\to frames};

\node[stage, text width=1.15cm] (h_face) at (5.05, 1.32) {MLP\\(pos-weighted)};
\node[stage, text width=1.15cm] (h_audio) at (5.05, 0.00) {MLP\\(pos-weighted)};
\node[stage, text width=1.15cm] (h_text) at (5.05, -1.32) {MLP\\(pos-weighted)};

\node[stage, text width=1.05cm] (p_face) at (6.6, 1.32) {$p^{face}(t)$};
\node[stage, text width=1.05cm] (p_audio) at (6.6, 0.00) {$p^{audio}(t)$};
\node[stage, text width=1.05cm] (p_text) at (6.6, -1.32) {$p^{text}(t)$};

\node[stage, text width=0.75cm] (fuse) at (8.25, 0.00) {Late\\fusion\\$p_t$};
\node[stage, text width=1.2cm] (smooth) at (9.65, 0.00) {Temporal\\smoothing\\$\delta$};
\node[stage, text width=1.10cm] (agg) at (11.20, 0.00) {Mean or\\max agg\\$\rightarrow s(v)$};
\node[stage, text width=1.00cm] (gate) at (12.65, 0.00) {Global\\text\\classifier};
\node[outbox, text width=0.62cm] (out) at (13.85, 0.00) {$\hat{y}(v)$};

\node[stage, text width=2cm] (global) at (9, -2)
  {Global text\\logistic\\regression};

\draw[arr] (video.east) -- ++(0.20, 0) coordinate (split);
\draw[arr] (split) |- (face.west);
\draw[arr] (split) |- (audio.west);
\draw[arr] (split) |- (text.west);

\draw[arr] (face.east) -- (h_face.west);
\draw[arr] (audio.east) -- (interp_a.west);
\draw[arr] (text.east) -- (interp_t.west);

\draw[arr] (interp_a.east) -- (h_audio.west);
\draw[arr] (interp_t.east) -- (h_text.west);

\draw[arr] (h_face.east) -- (p_face.west);
\draw[arr] (h_audio.east) -- (p_audio.west);
\draw[arr] (h_text.east) -- (p_text.west);

\draw[arr] (p_audio.east) -| ++(0.20, 0) coordinate (split2);
\draw[arr] (p_face.east) -| (split2);
\draw[arr] (p_text.east) -| (split2);
\draw[arr] (split2) -- (fuse.west);

\draw[arr] (fuse.east) -- (smooth.west);
\draw[arr] (smooth.east) -- (agg.west);
\draw[arr] (agg.east) -- (gate.west);
\draw[arr] (gate.east) -- (out.west);

\draw[arr] (video.south) -- ++(0, -1.35) -| (global.west);
\draw[arr] (global.east) -| (gate.south);

\end{tikzpicture}}
\caption{Proposed A/H pipeline: frozen face, audio, and text descriptors; frame-level MLPs; late fusion; temporal aggregation; optional global-text gate for video-level Macro F1.}
\label{fig:ah_pipeline}
\end{figure}
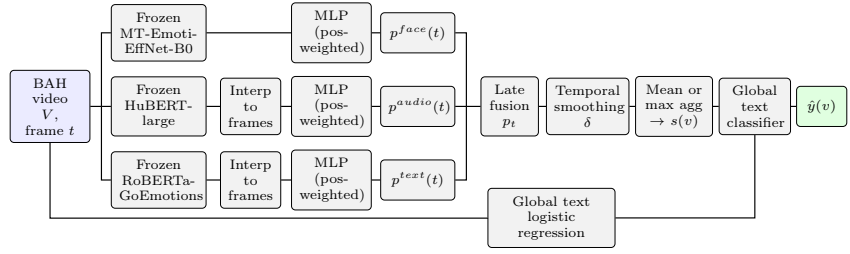

\subsubsection{Multimodal feature extraction}
\textbf{Visual modality.} For each frame, we use MT-EmotiEffNet-B0~\cite{savchenko2023hse}, pre-trained on VGGFace2~\cite{cao2018vggface2} and fine-tuned on AffectNet, to extract a $1280$-dimensional facial embedding $\mathbf{x}^{face}(t)$ and ten expression/VA scores $\mathbf{s}(t)$ from the official cropped-aligned faces.

\textbf{Audio modality.} The audio track of each video is converted to $16$\,kHz mono wav and processed by HuBERT-large~\cite{hsu2021hubert}, producing $1024$-dimensional hidden states $\mathbf{h}^{audio}(\tau)$ along the acoustic timeline.

\textbf{Text modality.} The official transcript of each video is encoded with RoBERTa-base trained on GoEmotions~\cite{demszky2020goemotions,liu2019roberta}, yielding a sequence of $768$-dimensional token/sentence embeddings $\mathbf{e}^{text}(\tau)$.

Since audio and text are not aligned with face frames, we linearly interpolate $\mathbf{h}^{audio}$ and $\mathbf{e}^{text}$ to the visual frame indices of each video, obtaining $\mathbf{h}^{audio}(t)$ and $\mathbf{e}^{text}(t)$.

\subsubsection{Frame-level classifiers and late fusion}
For each modality we train a separate feed-forward network (one hidden layer with $64$ ReLU units and a single logistic output) on frame-level BAH annotations. Training uses binary cross-entropy with logits and a positive-class weight $w_{+}=N_{neg}/N_{pos}$ to compensate for class imbalance; all frames labeled as A/H are kept, while negative frames are subsampled with step $1$. For the final submission, each modality classifier is fine-tuned on the concatenation of the official training and validation splits. This yields frame-wise probabilities $p^{face}(t)$, $p^{audio}(t)$, and $p^{text}(t)$.

The multimodal frame score is obtained by late fusion:
\begin{equation}\label{eq:ah_fusion}
p_t = w_1\, p^{face}_t + w_2\, p^{audio}_t + w_3\, p^{text}_t,
\qquad w_1+w_2+w_3=1.
\end{equation}
 We also study early fusion, where $[\mathbf{x}^{face}(t),\mathbf{h}^{audio}(t),\mathbf{e}^{text}(t)]$ is concatenated and fed into one MLP, but the late-fusion ensemble performs better in our experiments.

\subsubsection{Video-level decision}
ABAW-11 requires one binary label per video. Starting from fused frame probabilities, we optionally apply temporal smoothing inside each video with half-width $\delta$:
\begin{equation}\label{eq:ah_smooth}
\tilde{p}_t=\frac{1}{|\mathcal{N}_\delta(t)|}\sum_{t_i\in\mathcal{N}_\delta(t)} p_{t_i},
\end{equation}
 where $\mathcal{N}_\delta(t)$ contains frame indices from the same video within $\delta$ frames of $t$. The video score is then computed by aggregating frame probabilities:
\begin{equation}\label{eq:ah_video_score}
s(v)=\mathrm{Agg}\bigl(\{\tilde{p}_t : t\in\mathcal{F}_v\}\bigr),
\end{equation}
 where $\mathcal{F}_v$ is the set of labeled frames in video $v$, and $\mathrm{Agg}$ is either the arithmetic mean or the maximum.

Following ABAW-8 solution~\cite{savchenko2025abaw8}, we additionally train a \textbf{global text classifier} at the video level. Pooled transcript embeddings (mean and standard deviation over tokens) are standardized and passed to logistic regression, producing a video-level text probability $p^{global}_{text}(v)$. A hard gate filters videos for which the transcript alone does not suggest A/H:
\begin{equation}\label{eq:ah_gate}
\hat{s}(v)=
\begin{cases}
0, & \text{if } p^{global}_{text}(v)<\tau_{global},\\
s(v), & \text{otherwise}.
\end{cases}
\end{equation}
 The final video label is
\begin{equation}\label{eq:ah_decision}
\hat{y}(v)=
\begin{cases}
1, & \text{if } \hat{s}(v)\ge\tau,\\
0, & \text{otherwise}.
\end{cases}
\end{equation}
Fusion weights $(w_1,w_2,w_3)$, smoothing width $\delta$, aggregation rule $\mathrm{Agg}$, thresholds $\tau$ and $\tau_{global}$, gate mode, and global-text hyperparameters are selected by grid search on the labeled public test split ($525$ videos). The global text classifier is trained on the concatenation of training and validation splits ($902$ videos).

\section{Experiments}\label{sec:exper}

\subsection{MTL Challenge}
\label{sec:exper_mtl}

We evaluate the ABAW-11 MTL pipeline (Sec.~\ref{sec:methods_mtl}) on the official training and validation splits of s-Aff-Wild2~\cite{kollias2022abaw}. Due to missing labels, $142{,}333$ training frames provide only $103{,}917$ valence/arousal values, $90{,}645$ expression labels, and $103{,}316$ AU annotations. The validation set contains $26{,}876$ faces with complete VA and AU labels, but only $15{,}440$ expression labels.

\subsubsection{Comparison with published methods}

\begin{table}[t]
\caption{Comparison with MTL scores on s-Aff-Wild2 from prior ABAW challenge publications.}
\label{table:mtl_sota}
  \centering
  {\begin{tabular}{@{}llcccc@{}}
\toprule
 Method  & $P_{VA}$ & $P_{EXPR}$ & $P_{AU}$ & $P_{MTL}$  \\
\midrule
Organizer baseline~\cite{kollias2023abaw}   & 0.12 & 0.10 & 0.12 & 0.34 \\
HSEmotion~\cite{savchenko2024leveraging}   & 0.56 & 0.42 & 0.52 & 1.49 \\
AIWELL-UOC~\cite{cabacas2024enhancing}   & 0.37 & 0.28 & 0.47 & 1.11 \\
SCU ACers~\cite{li2024affective}  & 0.37 & 0.30 & 0.49 & 1.16 \\
HFUT-MAC1~\cite{shen2024facial}  & 0.38 & 0.30 & 0.50 & 1.18 \\
HSEmotion~\cite{savchenko2024leveraging}  & 0.41 & 0.33 & 0.51 & 1.25 \\
Netease Fuxi AI Lab~\cite{liu2024affective}  & 0.54 & 0.43 & 0.56 & 1.53 \\
\midrule
ABAW-11 baseline~\cite{liu2022convnext,kollias2026abaw11}  & -- & -- & -- & 0.45 \\
Ours & 0.56 & 0.46 & 0.54 & \textbf{1.56} \\
\bottomrule
\end{tabular}}
\end{table}

Table~\ref{table:mtl_sota} summarizes MTL scores reported in prior ABAW publications on s-Aff-Wild2. Ablation of our ABAW-11 pipeline on the labeled validation split is given in Tables~\ref{table:mtl_ablation} and~\ref{table:mtl_pp_ablation}.

On the s-Aff-Wild2 validation split, our ensemble reaches $P_{MTL}=1.56$, improving over the ABAW-7 validation ensemble ($1.49$, same split protocol) and over the official ABAW-11 ConvNeXt baseline ($0.45$).

\subsubsection{Ablation of the proposed pipeline}
Table~\ref{table:mtl_ablation} isolates the contribution of each component of the proposed ABAW-11 MTL pipeline. Frame-level MT-EmotiDDAMFN and MT-EmotiEffNet-B0 heads reach $P_{MTL}=1.31$ and $1.27$, respectively. Post-processing improves DDAMFN to $1.53$ and EffNet-B0 to $1.49$; weighted backbone blending further raises the score to $1.56$.

\begin{table}[t]
\caption{Ablation of the proposed ABAW-11 MTL pipeline on the validation set.}
\label{table:mtl_ablation}
  \centering
  \begin{tabular}{@{}lllll@{}}
\toprule
 Configuration & $P_{VA}$ & $P_{EXPR}$ & $P_{AU}$ & $P_{MTL}$  \\
\midrule
MT-EmotiDDAMFN + heads & 0.48 & 0.33 & 0.50 & 1.31 \\
MT-EmotiEffNet-B0 + heads & 0.44 & 0.34 & 0.49 & 1.27 \\
MT-EmotiDDAMFN + post-processing & 0.56 & 0.44 & 0.52 & 1.53 \\
MT-EmotiEffNet-B0 + post-processing & 0.52 & 0.45 & 0.52 & 1.49 \\
Ensemble (DDAMFN + EffNet) & 0.56 & 0.46 & 0.54 & 1.56 \\
\bottomrule
\end{tabular}
\end{table}

Table~\ref{table:mtl_pp_ablation} details the incremental post-processing steps for MT-EmotiDDAMFN on a single backbone. VA smoothing provides the main VA gain; temporal EXPR smoothing and per-class bias improve expression recognition; AffectNet blending adds a further EXPR gain in the single-backbone setting; per-AU threshold tuning adds the final AU boost. In the final ensemble, AffectNet blending is applied after weighted backbone blending (Fig.~\ref{fig:mtl_pipeline}). A grid search over $\sigma$ and $\delta$ (Eq.~\ref{eq:gauss}) shows that smoothing helps VA and EXPR (up to $+0.08$ and $+0.11$ for DDAMFN) but not AU detection. Interestingly, temporal smoothing contributes nearly all improvements in valence-arousal prediction, whereas expression calibration mainly benefits categorical recognition. AU detection benefits almost exclusively from threshold optimization, indicating that calibration requirements differ substantially across affective tasks.

\begin{table}[t]
\caption{Incremental post-processing ablation for MT-EmotiDDAMFN.}
\label{table:mtl_pp_ablation}
  \centering
  \begin{tabular}{@{}lcccc@{}}
\toprule
 Step & $P_{VA}$ & $P_{EXPR}$ & $P_{AU}$ & $P_{MTL}$  \\
\midrule
Frame-level heads & 0.48 & 0.33 & 0.50 & 1.31 \\
+ VA smoothing & 0.56 & 0.33 & 0.50 & 1.39 \\
+ EXPR smoothing & 0.56 & 0.38 & 0.50 & 1.44 \\
+ EXPR bias & 0.56 & 0.43 & 0.50 & 1.49 \\
+ AffectNet blending & 0.56 & 0.44 & 0.50 & 1.50 \\
+ AU threshold tuning & 0.56 & 0.44 & 0.52 & 1.53 \\
\bottomrule
\end{tabular}
\end{table}

The final evaluation protocol applies the validation-selected post-processing without further adaptation and ensemble weights are applied to $51{,}159$ official test frames using heads trained on the training split only.

\subsection{A/H Video Recognition Challenge}
\label{sec:exper_ah}

We evaluate the ABAW-11 A/H pipeline (Sec.~\ref{sec:methods_ah}) on the official BAH splits~\cite{kollias2025bah,kollias2026abaw11}: $778$ training videos, $124$ validation videos, and $525$ public test videos. 

\subsubsection{Comparison with published methods}
\begin{table}[t]
\caption{Reported video-level Macro F1 on BAH from prior ABAW challenge publications.}
\label{table:ah_published}
  \centering
  \begin{tabular}{@{}llc@{}}
\toprule
 Method  & Macro F1 \\
\midrule
ABAW-10 baseline~\cite{kollias2026abaw10}  & 0.343 \\
Lenovo PCIE~\cite{kollias2026abaw10}   & 0.675 \\
LEYA~\cite{ryumina2026leya}  & 0.714 \\
Fennec~\cite{bekhouche2026conflictaware}  & 0.715 \\
VisPBF~\cite{pereira2026brother}  & 0.727 \\
\hline
ABAW-11 baseline (Video-LLaVA zero-shot)~\cite{kollias2026abaw11} & 0.283 \\
Ours & \textbf{0.731} \\
\bottomrule
\end{tabular}
\end{table}

Table~\ref{table:ah_published} lists video-level Macro F1 scores reported in prior BAH challenge publications. After the competition concludes, we will update the table with the final test-set results. Component and fusion ablations of our ABAW-11 pipeline are in Tables~\ref{table:ah_ablation_frame} and~\ref{table:ah_ablation_video}. On the ABAW-11 public test set, the proposed pipeline reaches video Macro F1 $0.73$, well above the official Video-LLaVA baseline ($0.283$).

\subsubsection{Ablation of the proposed pipeline}
Table~\ref{table:ah_ablation_frame} reports frame-level diagnostics on the ABAW-11 validation split ($79{,}538$ labeled frames). Text is the strongest unimodal cue (Weighted F1 $0.77$), and late fusion outperforms early fusion ($0.79$ vs.\ $0.76$).

\begin{table}[t]
\caption{Frame-level ablation of the proposed A/H pipeline on ABAW-11 validation ($\tau=0.5$ unless noted).}
\label{table:ah_ablation_frame}
  \centering
  \begin{tabular}{@{}lcc@{}}
\toprule
 Configuration & Weighted F1 & Macro F1 \\
\midrule
Face MLP only & 0.71 & 0.52 \\
Audio MLP only & 0.67 & 0.53 \\
Text MLP only & 0.77 & 0.59 \\
Early fusion MLP ($\tau=0.4$) & 0.76 & 0.60 \\
Late fusion ($\tau=0.55$) & 0.79 & 0.59 \\
\bottomrule
\end{tabular}
\end{table}

Table~\ref{table:ah_ablation_video} summarizes video-level ablations on ABAW-11 validation and the public test split. On validation, internal baselines (always-negative majority class, RandomForest, early fusion) are outperformed by late fusion; on the public split, max aggregation with the global-text gate yields the best Macro F1 ($0.731$). The best public configuration uses max aggregation, late fusion weights $(0.20,0.45,0.35)$, smoothing $\delta=40$, gate threshold $\tau_{global}=0.41$, and $\tau=0.35$. A transcript-only global classifier reaches a comparable video Macro F1 ($0.734$) with higher AP ($0.87$ vs.\ $0.82$), confirming that spoken language is a very strong cue; the hard gate restricts audiovisual fusion to videos whose transcript already suggests A/H.

\begin{table}[t]
\caption{Video-level ablation of the proposed A/H pipeline on ABAW-11 validation ($124$ videos) and the public test split ($525$ videos). The validation majority baseline always predicts no A/H.}
\label{table:ah_ablation_video}
  \centering
  {\begin{tabular}{@{}llcc@{}}
\toprule
 Configuration & Split & Macro F1 & AP \\
\midrule
Majority class & val & 0.28 & 0.60 \\
RandomForest on concatenated features & val & 0.67 & 0.82 \\
Early fusion MLP & val & 0.68 & 0.85 \\
Late fusion, mean aggregation & val & 0.72 & 0.85 \\
Late fusion, max aggregation & val & 0.71 & 0.83 \\
\hline
Late fusion, mean & public & 0.69 & 0.79 \\
Late fusion, mean + hard global-text gate & public & 0.71 & 0.80 \\
Late fusion, max & public & 0.69 & 0.82 \\
Late fusion, max + hard global-text gate & public & 0.73 & 0.82 \\
Global text classifier only & public & 0.73 & 0.87 \\
\bottomrule
\end{tabular}}
\end{table}

Frame MLPs are fine-tuned on all $902$ training and validation videos, the global text classifier is trained on the same $902$ videos, and fusion and gate hyperparameters are selected by grid search on the public test split. The primary submission applies the max+gate configuration to the private test set ($152$ videos).

\subsection{Discussion}
Our experiments reveal three consistent observations. First, temporal calibration contributes substantially more than replacing lightweight frozen backbones with larger architectures, suggesting that temporal consistency rather than feature quality is the dominant error source on s-Aff-Wild2. Second, transcript information remains the strongest modality for ambivalence recognition, while audiovisual cues mainly improve borderline examples. Finally, both challenges indicate that competitive affect recognition can be achieved without end-to-end fine-tuning, reducing computational cost while simplifying deployment.

\section{Conclusion}\label{sec:concl}

This paper advocates a \emph{compute-on-device, fuse-locally} design for in-the-wild affect analysis: strong facial, audio, and text encoders remain frozen, while accuracy is recovered through lightweight heads, validation-time post-processing, and modality-specific fusion rather than end-to-end fine-tuning of large backbones. For privacy-sensitive settings, raw video and transcripts need not leave the device; only compact frame-level scores or pooled embeddings are combined downstream (Figs.~\ref{fig:mtl_pipeline} and~\ref{fig:ah_pipeline}).

On s-Aff-Wild2 validation, post-processing and dual-backbone fusion raised $P_{MTL}$ from $1.31$ to $1.56$ (Tables~\ref{table:mtl_ablation}), outperforming both the official ConvNeXt baseline ($0.45$) and our ABAW-7 validation ensemble ($1.49$). The gain comes mainly from temporal smoothing, expression calibration, and per-AU thresholding applied \emph{after} frozen feature extraction, suggesting that much of the remaining MTL error on wild faces reflects temporal inconsistency and calibration rather than backbone capacity alone.

For A/H, the same frozen-extractor principle extends to video-level decisions: late fusion with a transcript gate reached Macro F1 $0.731$ on the public development split (Table~\ref{table:ah_ablation_video}), while a transcript-only global classifier reached $0.734$ with higher AP. Hesitancy is therefore largely lexical, but audiovisual fusion remains useful when the transcript is ambiguous. This supports a two-stage deployment pattern: a fast text screen at the video level, with multimodal fusion reserved for borderline cases.

The source code of training scripts, evaluation code, hyperparameter configurations, and inference pipelines for both challenges is publicly available at\footnote{\url{https://github.com/bakinalexey/abaw-11-mtl-bah-recognition}}. Overall, our results suggest that careful prediction calibration and multimodal fusion can compensate for much of the performance typically attributed to increasingly large backbone models. This observation encourages future research on efficient affective computing systems that prioritize robustness, interpretability, and deployment efficiency over backbone scale. The proposed framework is particularly suitable for privacy-sensitive applications such as mobile affect recognition, digital health, and human-computer interaction, where transmitting raw audiovisual data is undesirable. 



%
%
\bibliographystyle{splncs04}
\bibliography{main}
\end{document}